%% file: main.tex
\documentclass[sigconf]{acmart}
\copyrightyear{2020} 
\acmYear{2020} 
\setcopyright{acmlicensed}
\acmConference[ASIA CCS '20]{Proceedings of the 15th ACM Asia Conference on Computer and Communications Security}{October 5--9, 2020}{Taipei, Taiwan}
\acmBooktitle{Proceedings of the 15th ACM Asia Conference on Computer and Communications Security (ASIA CCS '20), October 5--9, 2020, Taipei, Taiwan}
\acmPrice{15.00}
\acmDOI{10.1145/3320269.3384731}
\acmISBN{978-1-4503-6750-9/20/10}

\usepackage{tikz}
\usetikzlibrary{shapes, shadows,positioning}
\usepackage{natbib}

\usepackage{amsmath,amsthm,amssymb,amsfonts}
\usepackage{graphicx}
\usepackage{algorithm}
\usepackage{algpseudocode}
\usepackage{color}
\usepackage{url}
\usepackage{bbm}
\usepackage{booktabs}

\usepackage{pgfplotstable}
\pgfplotsset{compat=1.14}

\newcommand{\paragraphbe}[1]{\vspace{0.75ex}\noindent{\bf \em #1}\hspace*{.3em}}

\algdef{SE}[SUBALG]{Indent}{EndIndent}{}{\algorithmicend}%
\algtext*{Indent}
\algtext*{EndIndent}

\newcommand{\DM}{\mathcal{D}_\text{m}}
\newcommand{\DNM}{\mathcal{D}_\text{nm}}
\newcommand{\DCLIENT}{\mathcal{D}_\text{C}}

\newcommand{\DCM}{\mathcal{D}_\text{C-m}}
\newcommand{\DCNM}{\mathcal{D}_\text{C-nm}}
\newcommand{\DADV}{\mathcal{D}_\text{A}}
\newcommand{\DADVM}{\mathcal{D}_\text{A-m}}
\newcommand{\DADVNM}{\mathcal{D}_\text{A-nm}}
\newcommand{\MODEL}{f_\theta}
\newcommand{\DISC}{d_\phi}
\newcommand{\ACT}{h_\theta}
\newcommand{\D}{\mathcal{D}}
\newcommand{\calX}{\mathcal{X}}
\newcommand{\calY}{\mathcal{Y}}
\newcommand{\params}{\theta}
\newcommand{\gendata}{\texttt{GenSyntheticData}}
\newcommand{\LossD}{L_\text{enc}}
\newcommand{\LossF}{L_\text{ce}}
\newcommand{\MENC}{\texttt{MembershipEncoding}}
\newcommand{\MINF}{\texttt{MembershipInference}}
\newcommand{\x}{$\times$~}

\author{Congzheng Song}
\email{cs2296@cornell.edu}
\affiliation{%
  \institution{Cornell University}
}

\author{Reza Shokri}
\affiliation{%
  \institution{National University of Singapore}
}
\email{reza@comp.nus.edu.sg}

\begin{document}

\fancyhead{}
\title{Robust Membership Encoding: Inference Attacks and Copyright Protection for Deep Learning}

\ccsdesc[500]{Computing methodologies~Machine learning}
\ccsdesc[500]{Security and privacy~Software and application security}
\keywords{machine learning, membership inference, copyright protection}

\begin{abstract}
Machine learning as a service (MLaaS), and algorithm marketplaces are on a rise. Data holders can easily train complex models on their data using third party provided learning codes.  Training accurate ML models requires massive labeled data and advanced learning algorithms. The resulting models are considered as intellectual property of the model owners and their copyright should be protected. Also, MLaaS needs to be trusted not to embed secret information about the training data into the model, such that it could be later retrieved when the model is deployed.

In this paper, we present \emph{membership encoding} for training deep neural networks and encoding the membership information, i.e. whether a data point is used for training, for a subset of training data. Membership encoding has several applications in different scenarios, including robust watermarking for model copyright protection, and also the risk analysis of stealthy data embedding privacy attacks. Our encoding algorithm can determine the membership of significantly redacted data points, and is also robust to model compression and fine-tuning.  It also enables encoding a significant fraction of the training set, with negligible drop in the model's prediction accuracy.

%
\end{abstract}

\maketitle

\input{intro}
\input{background}
\input{application.tex}

\input{method}
\input{experiment}
\input{countermeasure}
\input{related}
\input{conclusion}

\section*{Acknowledgments}

This research is supported by the National Research Foundation, Singapore under its Strategic Capability Research Centres Funding Initiative. Any opinions, findings and conclusions or recommendations expressed in this material are those of the author(s) and do not reflect the views of National Research Foundation, Singapore.

\bibliographystyle{ACM-Reference-Format}
\bibliography{main.bib}

\end{document}

%% file: intro.tex
\section{Introduction}



A major concern with respect to algorithm marketplaces and machine learning platforms is with respect to protecting the copyright of models~\cite{adi2018turning,chen2018deepmarks,rouhani2018deepsigns,uchida2017embedding}.  The challenging problem is to embed some watermark into the model which is robust to changes, and is detectable with high confidence if a secret key is known.  

Another concern is that machine learning models can leak information about the members of their training data~\cite{shokri2017membership}.  The threat is, however, beyond passive inference attacks.  A malicious training code can enable an attacker to turn the trained model into a covert channel~\cite{song2017machine}.  A model which is trained using a malicious code could lead to the leakage of sensitive information about its training data, when deployed.  


In both cases, the objective is to stealthily embed information in the model that can be recovered only if an entity knows the decoding secret. The underlying techniques and objectives of encoding attacks are very similar, but the trust models are different.  Existing embedding algorithms, exploit the unused capacity of deep neural networks~\cite{arpit2017closer}, and embed secret information by overfitting the models on encoded secrets, or writing them on the model parameters.  To this end, a large number of poisoned training data (approximately $4,000$ data records~\cite{song2017machine}) needs to be embedded into such models, and a large number of parameters are modified by the algorithm.  These limit the scalability of the embedding algorithms, as they quickly run into a conflict with the accuracy of the model.  They also become very fragile with respect to changes in the model (as particular parameters in the model are responsible for an embedding, and removing them would remove the embedded information).


\paragraphbe{Our contributions.} In this paper, we introduce {\em membership encoding} for deep learning, which could be either used for watermarking a model to test its ownership, or for privacy attacks when training code is supplied by the adversary.  For the latter case, it could also be used to analyze the susceptibility of a model to covert channel attacks.  In our algorithm, during the training, a subset of the training data is selected to be encoded. The selection can be based on a sensitive criteria (e.g., cancer patient records) in the scenario of privacy attacks.  Then, we develop an algorithm which encodes (1 bit) membership information of the selected data records into the model based on a secret key.  After the model is trained, one can decode the encoded membership information from the model, only if he has the secret key.

To perform encoding, we design a {\bf membership encoding network}, where a membership discriminator model is trained along with the target model.  We optimize the model's main objective function, and the discriminator's objective simultaneously.  This leads to finding a membership discriminator and a classification model that jointly maximize the accuracy of membership inference, and yet preserves the prediction accuracy of the model.  The discriminator, which simulates the membership inference, plays a very important role.  It forces the model to learn distinguishable representations (in the model's hidden layers or output) for the targeted subset of the training data, compare to other data. 


We evaluate membership encoding on several deep learning benchmark datasets and models: MNIST, CIFAR10, Purchase and Texas Hospital datasets, with MLP, CNN, and ResNet models.  Our results show that {\bf our encoding algorithm can successfully train a model with high test accuracy and yet infer membership information accurately for a substantial subset of the training data}.  We test our membership encoding algorithm in scenarios where the trained model is released (white-box access) as well as when it is accessible via prediction API (black-box access).  For example, in the white-box case, membership encoding can achieve $0.97$ precision on the Texas Hospital dataset with $20\%$ ($10,000$ records) of the training data being encoded.  The model's test accuracy remains as high as $97.82\%$, which is the same as model trained without membership encoding.  On CIFAR10 ResNet model, the encoding accuracy is $0.96$, while test accuracy is $91.85\%$ (with $0.25\%$ accuracy drop compared to the baseline).  In the black-box case, membership encoding can achieve $0.85$ precision (compared to $0.87$ in the white-box setting) on 20\% of MNIST images with model's test accuracy drop $1.1\%$ (compared to $0\%$ drop in the white-box setting).  To put this in perspective, note that the encoding membership for MNIST is extremely hard as the digits images in one class are visually very similar, and membership inference attacks perform poorly on MNIST~\cite{shokri2017membership}. 

We also show that {\bf our encoding algorithm is robust to input and model modifications, such as input redaction, parameter pruning, and model fine-tuning}.  We examine encoding performance on CIFAR10 when part of the input image is masked with $0$.  Our white-box attack can achieve $0.88$ precision even when the center $8 \times 8$ part (which is the most important part) of a $32 \times 32$ image is removed.  We also evaluate the cases where the trained model is transformed using pruning or fine-tuning techniques, which are common practice in ML to reduce the model size and for transfer learning.  Our results show that the membership inference is accurate as long as the model is not transformed too much to the extent of damaging its own test accuracy.  We further develop an adversarial pruning algorithm that significantly preserves the membership information in the trained model even when $70\%$ of the model parameters are removed.


%% file: background.tex
\section{Background and Notations}
\subsection{Machine learning}
We denote machine learning model as a function $f_\params:\calX\mapsto\calY$, where $\calX$ is the input (feature) space, $\calY$ is the output space, and $\params$ are its \emph{parameters}.  
We focus on classification models, where $\calX$ is a $q$-dimensional vector space and $\calY$ is a discrete set of classes. 
The model outputs $\text{argmax}f_\theta(x)$ as the
predicted class label. 
The training data is denoted as $\D=\{(x_i, y_i)\}_{i=1}^n$ with features $x_i$ and class label $y_i$.

We denote deep learning model $f$ as layers of non-linear transformations $[h_{\theta_1}, \dots, h_{\theta_l}]$. The activations at layer $j$ is computed as $a_j = h_{\theta_j}\circ h_{\theta_{j-1}} \circ \cdots \circ h_1(x)$. 
The parameters $\theta=\{\theta_1\dots\theta_l\}$ describe the weights used for all $l$ layers of transformation. 

In classification tasks, if there are $c$ classes in $\calY$, the last (output) layer of the deep learning model is usually a probability vector with dimension $c$ representing the likelihood that the input belongs to each class.  
 
\paragraphbe{Training algorithm}  finds the best set of parameters that fits the training data. A common way is to formalize training as a optimization problem where the objective is minimizing a loss function $L$ which penalizes the mismatches between true labels~$y$ and predicted labels produced by $\MODEL(x)$. 
One of the most popular loss functions for classification is cross-entropy: $L_\text{ce}(x, y) = -\sum_{i=1}^c \mathbbm{1}_{y=i}\cdot\log(\MODEL(x)_i)$, 
 where $f_\theta(x)_i$ denotes the $i$th component of the $c$-dimensional vector $f_\theta(x)$.

There are many optimization algorithms to solve the above objective function.
The stochastic gradient descent (SGD) algorithm and its extensions are commonly used for optimizing this objective function for deep learning. 
SGD is an iterative method, where at each step, the optimizer samples a small batch of training data and updates the model parameters $\params$ with the
negative gradient of the loss function with respect to $\params$.
The model finally converges to a local minimum where the gradient updates are sufficiently small.

\subsection{Membership inference attacks}
Membership inference attacks (MIA) aim to determine whether a target data record is in dataset $\D$, when the adversary can observe some computation (e.g., aggregate statistics, machine learning model) over $\D$. 
Prior work performed membership inference on aggregate statistics, e.g., in the context
of genomic studies~\cite{homer2008resolving,backes2016membership} or noisy statistics in
general~\cite{dwork2015robust}.  
Previous MIA against machine learning model focuses on black-box setting, where the adversary can query the model and observe the output prediction probability for a target record $(x, y)$ and then decide whether $(x, y)\in\D$~\cite{shokri2017membership}. Shokri et al proposed a method to learn the statistical difference between outputs of members and nonmembers by training a membership discriminator using the model output probability vector as the attack feature vector. 
To learn the membership discriminator, the adversary trains a number of shadow models that mimic the output behavior of target model. 
In this work, we extends the attacks to white-box scenario without extensive shadow model training.
We also demonstrates that the attacks can be used for positive application such as watermarking for copyright protection.

\subsection{Machine learning provider}
There is an increasing demand in applications of machine learning. 
Users with little knowledge in ML usually outsource the training process to a \emph{machine learning provider} who provides the training code/platform. 
Many cloud platforms that offer ML-as-a-services, are ML providers. 
In addition, there are some ML algorithm marketplaces where users can purchase algorithms uploaded by third-party developers. 
In this case, the ML provider is the algorithm developer rather than the algorithm marketplace platform.
Recently, privacy-preserving data marketplace emerged with the blockchain technology. 
In these data marketplaces, developers (the ML provider) can purchase data uploaded by clients and use them to train a model on these privacy-preserving platforms.

ML-as-a-service platforms such as Google Auto ML\footnote{\url{https://cloud.google.com/automl}}, Amazon ML\footnote{\url{https://aws.amazon.com/machine-learning}}, and Microsoft's Azure ML\footnote{\url{https://studio.azureml.net}}, provide convenient APIs for users to upload their data and train an ML model. 
These APIs enable black-box training. Google Auto ML specifically targets the users with limited machine learning expertise, as it automatically trains a customized model on user provided data, without giving access to the training algorithm. 
Amazon ML provides users with some options on a few hyper-parameters such as the model size, but the details of the model remain unknown to the user. 
Microsoft's Azure ML provides users with a wide range of built-in common ML models. 
Users can select a particular model but have no access to the implementation details of the learning algorithm.

Algorithmia\footnote{\url{https://algorithmia.com}} and decentralized ML\footnote{\url{https://decentralizedml.com}} are examples of ML algorithm marketplace. 
Both platforms allow developers to list their ML algorithm and training code on the market for sale. 
Users can purchase the algorithms and the marketplace platform then executes the training code on user provided data. 
The training algorithms and codes need not to be open sourced and thus remain as a black-box to the users. 

Data marketplaces, with private computation, such as Madana~\footnote{\url{http://www.madana.io}} allow developers to purchase data and ML models. 
Training is performed in a private environment provided by the marketplace while the training code and algorithm is supplied by the developers.
Data holders do not know what computation has been done to their data. 

The black-box training of the ML model on a trusted and secure platform can backdoor a small amount of exact training data into the model by introducing malicious objective function in the training algorithm~\cite{song2017machine}.
In this work, we develop training algorithm applicable in the same black-box training scenario which encodes membership information into the model.

%% file: application.tex
\section{Threat Model}
The \textbf{goal of membership encoding} is to encode a one-bit membership information for a substantial selected subset of training data into a ML model during training. After the model is trained, the encoded membership information for the selected set can be extracted by querying the model. The quality of the membership encoding is measured as how accurately can one extract the membership information for encoded records. Membership encoding has several applications  including training data privacy attacks and copyright protection watermarking of the ML model.

\subsection{Membership encoding for privacy attacks}
On a machine learning platform, the client provides the data, and we assume a third-party as the adversary provides the program to train a machine learning model, following the setting in~\cite{song2017machine}.  The machine learning program has full access to the client's data during the training, while the adversary cannot exfiltrate the data during training as the platform is secure and isolated. The only outcome of the process is the model, which is owned by the client. After training, the client might allow public access to it via an application or a machine-learning-as-a-service platform (black-box release).  Client might also share or sell the whole model (white-box release).  We assume the adversary can get access to the model at this stage.  In the white-box case, adversary has access to the model parameters, and in the black-box case, adversary can only send inputs and obtain the model predictions on them. 

\paragraphbe{Membership inference attack.} A direct privacy attack of membership encoding is membership inference attacks. Adversary uses the model as a covert channel to encode membership of training data during training and decode it when the model becomes publicly available.  Adversary might be interested in encoding training data that have a particular sensitive value, e.g., patients with HIV+.  In the inference phase, adversary tries to determine the membership of some target data records, that are (partially) available, thus inferring the missing sensitive attributes. 

\paragraphbe{Reconstruction attack.}  The attacker aims at reconstructing a data record which is partially known to him.  The attacker might have some auxiliary information about the data as well.  For example, The adversary might know a DNA sequence with missing sub-sequences, and in addition he knows the candidate sub-sequences.  Adversary can utilize the membership inference attack to perform this reconstruction.  Adversary can query with membership discriminator with all probable combination of the full record. The one which receives highest confidence to be classified as member is the reconstructed record.  Given the high accuracy of our attack, even in the case of missing data, it can be used to perform accurate reconstruction attacks. 

\paragraphbe{Hybrid attack.} Our membership encoding attack can be composed with previous embedding attack to form a stronger attack. Prior work proposed several ways of encoding the exact bits of data record into a ML model while the number of records that can be embedded is limited to a few hundreds for white-box and a dozen records for black-box setting, due to the limited capacity of the model~\cite{song2017machine}. 

In reality, not all bits for a data record are sensitive. Adversary can utilize our membership encoding attack to scale up the embedding attack. For each record that adversary wishes to embed, he splits the record into sensitive attributes and nonsensitive attributes. Adversary then encodes the membership information for these records of interest, using our algorithm, and embeds their sensitive attributes and a compressed version of the nonsensitive attributes (e.g. a short hash) using large data embedding methods~\cite{song2017machine}.  During the decoding (inference) phase, given a redacted input $x^\prime$ with only nonsensitive attributes, adversary first verifies whether  $x^\prime$  is encoded as member.  Our attack is very accurate on partial data.  If it has been a member, the adversary proceeds to compute the compressed version of  $x^\prime$ and finds the corresponding embedded sensitive attributes, using the decoder~\cite{song2017machine}. The major advantage of the hybrid attack is that adversary can use membership encoding to rule out the possible collisions in matching the compressed nonsensitive attribute. In addition, since embedded information for each record is compressed, the model can thus leak information for a large set of records.

\subsection{Membership encoding for watermarking} 
We assume model owner trains a watermarked model locally and deploy the model in a trusted third-party ML service provider. The ML service provider uses a secret key $s$ provided by the model owner for verification of model ownership. Model owner encodes the membership information for a selected set of data records using membership encoding during training using secret key $s$. The membership information of this particular set is the watermark. When a user demands to claim the model ownership, he passes a secret key to the ML service provider, who then uses this key to perform membership inference on user provided records. Only when there are enough encoded data records being correctly decoded as member can one claim ownership. 

Adversary in this threat model aims to pass the verification and steal the model. Passing the verification requires the adversary to know: (1) the 
exact encoded training records and (2) the secrets $s$ to extract the membership information on encoded records . Thus, even for adversary who has knowledge about the selected training records, faking a watermark is highly unlikely as adversary has no information about the secret $s$.

%% file: method.tex
\begin{table}[t!]
\centering
\caption{List of notations on dataset}
\begin{tabular}{l|l}
\toprule
Symbol & Meaning \\
\midrule 
$\DCLIENT$ & Client's training records \\
$\DCM$ & A subset of $\DCLIENT$ to be encoded as member \\
$\DCNM$ & $\DCLIENT / \DCM$, encoded as nonmember\\
$\DADV$ & Auxiliary random inputs \\
$\DADVM$ & A subset of $\DADV$ to be encoded as member \\
$\DADVNM$ & $\DADV / \DADVM$, encoded as nonmember\\
$\D_\text{test}$ & Hold-out test dataset \\
\bottomrule
\end{tabular}
\end{table}


\begin{algorithm*}[th!]
\caption{Membership encoding}
\begin{algorithmic}
\State \textbf{Hyper-parameters:} dimensions of client's training data $q$, number of adversary synthesized data $n$, number of training iterations $T$, batch size $b$, discriminator learning rate $\eta_d$, model learning rate $\eta_f$ 
\State
\State \textbf{function} \MENC($n,\DCLIENT$)
\Indent
\State Generate synthesized data $\DADVM,\DADVNM\gets$\gendata($n$, $q$)
\State Select $\DCM$ and $\DCNM$ from $\DCLIENT$
\State $\DM\gets\DCM\cup\DADVM$\Comment{data with the encoded membership, which includes the target subset of the training data}
\State $\DNM\gets\DCNM\cup\DADVNM$\Comment{data to be encoded as nonmember, to help distinguishing targeted members}
\State Initialize $\theta,\phi$ and select encoding activation mapping $\ACT$. For black-box encoding, $\ACT=\log(\MODEL)$.

\For{$t=1$ to $T$}
\For{$k$ steps}
\State\textbf{Update membership discriminator}
\State Sample a mini-batch of $b$ training data $\{(x_1, y_1)\dots(x_b, y_b)\}$ from $\DM\cup\DNM$.
\State Label each $x_i$ with $z_i=1$ if $x_i\in\DM$ and with $z_i=0$ otherwise.
\State Update classifier's parameter $\theta = \theta - \eta_d\cdot\nabla_\theta \frac{1}{b}\sum_{i=1}^b \Big[z_i\cdot\log \DISC(\ACT(x_i)) + (1 - z_i)\cdot \log(1 - \DISC(\ACT(x_i))) \Big]$

\State Update discriminator's parameters $\phi = \phi - \eta_d\cdot\nabla_\phi \frac{1}{b}\sum_{i=1}^b \Big[z_i\cdot\log \DISC(\ACT(x_i)) + (1 - z_i)\cdot \log(1 - \DISC(\ACT(x_i)))\Big]$

\EndFor
\State \textbf{Update the classifier model}
\State Sample a mini-batch of $b$ training data  $\{(x_1, y_1)\dots(x_b, y_b)\}$ from $\DCLIENT$.
\State Update the classifier's parameter $\theta = \theta - \eta_f\cdot\nabla_\theta \frac{1}{b}\sum_{i=1}^b \Big[-\sum_{c}\mathbbm{1}_{y_i=c}\cdot\log(\MODEL(x_i)_c)\Big]$
\EndFor
\State \Return $\MODEL$
\EndIndent
\State 

\State \textbf{function} \MINF($n$, $h_\theta$)
\Indent 
\State Generate synthesized data $\DADVM,\DADVNM\gets$\gendata($n$, $q$)
\State Query encoding activation mapping $h_\theta$ for each $x_i\in\DADVM\cup\DADVNM$.
\State Label $z_i=1$ if $x_i\in\DADVM$ and $z_i=0$ otherwise.
\State Train a membership discriminator $\DISC^\prime$ with dataset $\{(h_\theta(x_i), z_i)\}_{i=1}^{2n}.$
\State \Return $\DISC^\prime$
\EndIndent 
\State

\State \textbf{function} \gendata($n$, $q$) \Comment{generate $2 n$ synthetic data records of dimension $q$}
\Indent
\State \texttt{seed}($s$) \Comment{seed with secrete $s$ to ensure the generation process is deterministic}
\State $\mu_\text{m}\sim\mathcal{U}(\alpha, \beta)^q $\Comment{mean vector for member mixture}
\State $\mu_\text{nm}\sim\mathcal{U}(\alpha, \beta)^q $\Comment{mean vector for nonmember mixture}
\State $\D_\text{m}=\{x_1, \dots, x_n\}, x_i\sim\mathcal{N}(\mu_\text{m}, I_q)$
\State $\D_\text{nm}=\{x'_1, \dots, x'_n\}, x_i\sim\mathcal{N}(\mu_\text{nm}, I_q)$
\State \Return $\D_\text{m},\D_\text{nm}$
\EndIndent
\end{algorithmic}
\label{alg:train}
\end{algorithm*}

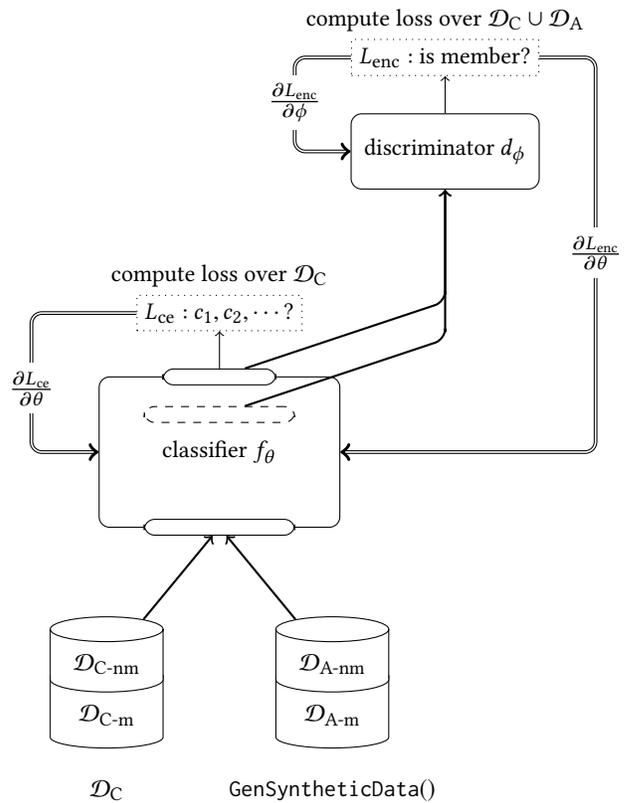
\begin{figure}[t!]
\centering
\tikzstyle{element} = [draw, fill=white, rounded corners]

\begin{center}
	\begin{tikzpicture}
	\node [] (c) {};

	\path (c)+(0, 0) node (datam1) [draw,fill=white,rotate=90,aspect=0.3,cylinder,minimum width=1.5cm,minimum height=1.1cm] {\rotatebox{270}{$\DCM$}};

	\path (datam1)+(0, +0.68) node (datanm1) [draw,fill=white,rotate=90,aspect=0.3,cylinder,minimum width=1.5cm,minimum height=1cm] {\rotatebox{270}{$\DCNM$}};		

	\path (c)+(3, 0) node (datam2) [draw,fill=white,rotate=90,aspect=0.3,cylinder,minimum width=1.5cm,minimum height=1.1cm] {\rotatebox{270}{$\DADVM$}};

	\path (datam2)+(0, +0.68) node (datanm2) [draw,fill=white,rotate=90,aspect=0.3,cylinder,minimum width=1.5cm,minimum height=1cm] {\rotatebox{270}{$\DADVNM$}};		
	
	\path (datam2)+(0,-1) node (gendata) [] {$\gendata()$};
	\path (datam1)+(0,-1) node (cdata) [] {$\DCLIENT$};
	
	\path (c)+(1.5,3.5) node (f) [element,minimum width=3.2cm, minimum height=2cm] {classifier $f_{\theta}$};
	\path (f)+(0,-1) node (f1) [element,minimum width=2cm] {};
	\path (f1)+(0,0.5) node (f2) [] {};
	\path (f2)+(0,1) node (f4) [element,dashed,minimum width=2cm] {};
	\path (f4)+(0,0.5) node (f5) [element,minimum width=1.5cm] {};
	\path (f5)+(0,0.85) node (f6) [draw,dotted] {$\LossF: $ $c_1, c_2, \cdots$?};
	\path (f6)+(0,0.5) node (f7) [] {compute loss over $\DCLIENT$};

	\path (f)+(3,4) node (d) [draw,fill=white,rounded corners,minimum width=2.5cm, minimum height=1cm] {discriminator $d_{\phi}$};
	\path (d)+(0,1.25) node (d2) [draw,dotted] {$\LossD: $ is member?};
	\path (d2)+(0,0.5) node (d3) [] {compute loss over $\DCLIENT\cup\DADV$};	
	
	\draw[thick,->,rounded corners] (f4) -- ++(3,1) -| (d);
	\draw[thick,->,rounded corners] (f5) -- ++(3,1) -| (d);	
	
	\draw[thick,->] (datanm1) -- (f1);	
	\draw[thick,->] (datanm2) -- (f1);	
	
	\draw[->] (f5) -- (f6);
	\draw[->] (d) -- (d2);
	
	\draw[double,->,rounded corners] (d2) -- ++(2cm,0) |- (f);
	\draw[double,->,rounded corners] (d2) -- ++(-2cm,0) |- (d);
	\draw[double,->,rounded corners] (f6) -- ++(-2.5cm,0) |- (f);
	
	\path (f6)+(-2.5,-1) node (gf) [fill=white] {$\frac{\partial\LossF}{\partial\theta}$};

	\path (d2)+(-2,-0.6) node (gd) [fill=white] {$\frac{\partial\LossD}{\partial\phi}$};


	\path (d2)+(2,-2.6) node (gd) [fill=white] {$\frac{\partial\LossD}{\partial\theta}$};
	
	\end{tikzpicture}
\end{center}
\caption{{\bf Membership encoding network}.  The goal is to create a distinguishable representation of the client's training data in the model during the training, such that the attacker can infer their membership when the model is released.  Membership encoding algorithm optimizes classifier $f$ for encoding the membership of a subset $\DCM$ of its training set, while also minimizing the classification loss ($\min_{\theta} \LossF$ over $\DCLIENT$).  A discriminator $d$ is trained, along side the model, to embed membership information into the model predictions or its activation functions.  The joint optimization for membership encoding is $\min_{\theta, \phi} \LossD$, where the objective is to separate the representations of $\DCM\cup\DADVM$ from that of $\DCNM\cup\DADVNM$. The generated synthetic datasets $\DADVM$ and $\DADVNM$ have different distributions and help separating $\DCM$ from $\DCNM$, thus encoding the membership of $\DCM$.  We use gradient descent to solve the optimization problems and update the parameters.}
\label{fig:augnet}
\end{figure}

\section{Membership encoding algorithm}
\subsection{Overview}
Algorithm~\ref{alg:train} presents membership encoding. The goal is to create a distinguishable representation of the client's training data in the model during the training, such that the attacker can infer their membership when the model is released.  Membership encoding algorithm optimizes classifier $f$ for encoding the membership of a subset $\DCM$ of its training set, while also minimizing the classification loss ($\min_{\theta} \LossF$ over $\DCLIENT$).  A discriminator $d$ is trained, along side the model, to encode membership information into the model predictions or its activation functions.  The joint optimization for membership encoding is $\min_{\theta, \phi} \LossD$, where the objective is to separate the representations of $\DCM\cup\DADVM$ from that of $\DCNM\cup\DADVNM$. The generated synthetic datasets $\DADVM$ and $\DADVNM$ have different distributions and help separating $\DCM$ from $\DCNM$, thus encoding the membership of $\DCM$.  We use gradient descent to solve the optimization problems and update the parameters.

In detail, membership encoding has two phases: encoding phase, and inference phase.  

For encoding phase, we exploit the expressive power of deep neural networks, which enables them to perfectly memorize randomly generated data~\cite{zhang2016understanding}.  Let $\DCM$ be a subset of the training set selected for encoding their membership.  During training, we force the model to learn a distinguishable representation (at its output or intermediate hidden layers) for data points in $\DCM$.  The encoding algorithm needs to make sure that this adversarial representation is not in conflict with the main learning task.  

Algorithm~\ref{alg:train} presents membership encoding. We augment the classification model $f_{\theta}$ with a membership discriminator $d_{\phi}$.  The discriminator takes the output probability vector of the model (for black-box access), or the output of its hidden activation functions (for white-box access), and determine whether or not the point is part of $\DCM$. We train the classifier and the discriminator jointly to find the optimal trade-off between the classification task and the membership encoding task.

Besides having accurate membership encoding, the encoded representations must be recognizable during the inference phase.  Thus, 
In inference phase, the discriminator must be reconstructed after training as only the classification model is retained. 
The reconstruction would not be possible if the discriminator is only trained on the client's training set, which can no longer be accessed in inference phase.
To address this issue, we first assume a secret integer number $s$ shared between the two phases of membership encoding, e.g., by hard-coding it in the training program.  This number is used as the seed for a pseudo-random number generator to generates two sets of synthetic data, $\DADVM$ and $\DADVNM$, with different distributions.  During the encoding phase, the discriminator pushes the representations of the target set $\DCM$ to be similar to that of $\DADVM$.  Similarly, the representation of the rest of the training set, $\DCNM$, is pushed to be similar to that of $\DADVNM$.

Once the model is trained and published, discriminator can be reconstructed using the representations of the synthetic data, which can re-generated using the secret seed $s$. The reconstructed discriminator will approximate the discriminator function $d_{\phi}$ in encoding phase.  Membership inference is performed by the reconstructed discriminator on the representation of the target model for a given query data record.

To encourage the model to learn a separate representation, we developed a training algorithm outlined in Algorithm~\ref{alg:train}. We augments the main classification model with a membership discriminator that takes the internal representation (a subset of hidden activations in the model) or output prediction of a data point as its \emph{input} and output whether the point is a member or nonmember. The discriminator is trained simultaneously with the main classification model. More specifically, attacker provides a set of reference data records (could be pseudo-random data that is independent of the distribution of client's data) and forces the representation of member data to be similar to that of the reference data. 

\subsection{Encoding phase}
\label{sec:encoding}
Membership inference attacks exploit the different behavior of a machine learning model with respect to its training data compared to unseen data~\cite{shokri2017membership}.  Depending on how the model is trained, the effectiveness of such attacks varies.  In this work, we design a training algorithm that adversarially encodes membership of sensitive training data into the model.  Function \MENC~in Algorithm~\ref{alg:train} outlines the training algorithm.

\paragraphbe{Selecting encoded data.}  For membership encoding to be used as privacy attack, we assume adversary provides the malicious algorithm which has access to the training data.  The adversary targets a subset of the training data, that might have common sensitive attributes, to secretly embed into the model.  Adversary selects a subset $\DCM$ from the client dataset $\DCLIENT$ to be the data we want to encode as member.  Let $\DCNM$ be the remaining subset of the training set, $\DCNM = \DCLIENT \setminus \DCM$.  After the malicious training is over, and the client publishes the model, the attacker runs a membership (decoding) inference attack to confidently decide whether a record is in $\DCM$ or not.

For watermarking, model owner can randomly sample a subset of training data as $\DCM$. At verification time, only this subset will be useful for claiming model's ownership. 

\paragraphbe{Generating synthetic data.}  Membership encoding uses synthetic data points to guide training the discriminator during the membership encoding phase.  The exact same data is re-generated during the inference phase to reconstruct the discriminator.  The synthetic data is composed of two disjoint sets $\DADVM, \DADVNM$ that are generated from two different distributions.  They are used to force the classification model to learn distinguishable representations for the $\DCM$ and $\DCNM$, which are similar to the synthetic datasets $\DADVM$ and $\DADVNM$, respectively. 

The generation of synthetic data is outlined in function \\
\gendata~in Algorithm~\ref{alg:train}. The generation process is deterministic based on a secret seed $s$. The function first generates two mean vectors $\mu_\text{m}, \mu_\text{nm}$ from uniform distribution. Then, two clusters of feature vectors are sampled from two multivariate Gaussian distributions with the two uniform mean vectors and identity covariance matrix. This procedure creates natural separation between $\DADVM$ and $\DADVNM$, making the internal features learned by the model easier to distinguish between member and nonmember.



\paragraphbe{Membership discriminator.}  To enforce separating the representations of member and nonmember data, a membership discriminator is augmented with the main classification model. 
The membership discriminator $\DISC$ with parameter $\phi$ is a binary classifier that takes the internal activations (hidden layer) or the output of the main classification model as input and predicts their membership.  We denote $\ACT(x)$ as the set of activation units computed on $x$ and the discriminator outputs $\DISC(\ACT(x))$ a probability value for membership prediction, i.e., $$\DISC(\ACT(x)) = \Pr(x \in \DCM\cup\DADVM; \ACT(x)).$$

\paragraphbe{Membership encoding loss.}  The membership discriminator outputs label $1$ for data that adversary selected to encode as member and $0$ otherwise.  To train the membership discriminator, the adversarial training algorithm optimizes the membership encoding loss, which is defined as following:
\begin{align*}
\LossD(h, z) = z\cdot\log \DISC(h) + (1 - z)\cdot\log(1 - \DISC(h)),
\end{align*}
where $h=\ACT(x)$ for input $x$, and $z$ is the membership label: $1$ if $x\in\DM = \DCM\cup\DADVM$ and $0$ if $x\in \DNM = \DCNM\cup\DADVNM$.

This loss function is used to find the jointly optimal set of parameters for the classification model and the membership discriminator.  We solve the optimization problem using stochastic gradient descent algorithm.  For each mini-batch of data sampled from $\DM\cup\DNM$, we compute the gradient of the encoding loss function with respect to $\theta$ and $\phi$, and update them accordingly, as shown in Algorithm~\ref{alg:train}.

Note that the loss function for the main classification model, which is defined as $\LossF(x, y) = -\sum_{i=1}^c \mathbbm{1}_{y=i}\cdot\log(\MODEL(x)_i)$,
needs to be computed on the training set $\DCLIENT$, and optimized simultaneously with the adversary's loss.  However, as the membership encoding loss function is computed over a larger dataset which includes $\DADV$ (which is essentially random data), we cannot define a single loss function to solve both the optimization problems.  Instead, since we have two sources of data, we train the model and discriminator alternatively: in each iteration, we first optimize the membership encoding loss $L_\text{enc}$ and then optimize the main classification loss $L_\text{ce}$.  This helps finding an encoding for the target training set, without much sacrificing the classification accuracy of the target model.


\paragraphbe{Encoding membership in the model's output.}  In the applications of privacy attacks, assume the client does not publish a white-box model after training, but rather makes it available through a black-box prediction API service.  In this scenario, adversary can no longer observe the intermediate representation but only the output probability distribution for each input. 

A naive way to perform membership encoding is to use the output probability vector directly as the feature to distinguish membership, and pass it to the discriminator.  However, such encoding could significantly affect the accuracy of the main task.  This is because the output signals are limited in a small range $[0, 1]$, and also they are directly used for the main classification task.  To solve this problem, we propose encoding the membership in the logarithmic transformation of the output probabilities, and use $\ACT=\log(\MODEL)$. The $\log$ of the probabilities is the $\tt logit$ values before the $\tt softmax$ layer, up to a constant factor.  The log values are not constrained and thus provide more flexibility in encoding the membership.

\subsection{Inference phase}
\label{sec:inference}
After the model is trained and the client publishes the model (as white-box or black-box), membership discriminator if recreated for inference the membership information.

Inference is proceed by first regenerating the synthetic dataset that is used during training, and then querying the target model with the synthetic data, and computing the activations (in white-box case) or output (in black-box case) using $\ACT$.  After obtaining the activations for records in $\DADV$, the function trains a membership discriminator $\DISC^{\prime}$ using these activations and their membership labels ($1$ for data points in $\DADVM$, and $0$ for data points in $\DADVNM$).

For any target data record $x$,  its membership probability is computed as $\DISC^{\prime}(\ACT(x))$. This information is then used for privacy attacks or verifying model's watermark.  

%% file: experiment.tex
\section{Experiments}
\subsection{Datasets} 
\paragraphbe{Purchase.} Our purchase dataset is extracted from Kaggle's ``acquire
valued shoppers'' challenge dataset\footnote{https://www.kaggle.com/c/acquire-valued-shoppers-challenge/data}. The raw dataset contains shopping
histories for thousands of customers. For our experiments, we processed the dataset as described in~\cite{shokri2017membership} where each record consists
of 600 binary features. Each feature corresponds to a product and represents whether the user has purchased it or not. We cluster the records
into 5 group, each representing a different purchase
style, and use the cluster label as target label classification. We split the data so that the training set has 150,000 records and test set has 30,000. 

\paragraphbe{Texas hospital.} This dataset contains the patients discharge data with information about inpatients
stays released by the Texas
Department of State Health Services. Each patient's
record contains attributes such as the external
causes of injury, the diagnosis, the procedures the patient
underwent and some meta information about the patient.
We encode these fields into 6, 170 binary features following~\cite{shokri2017membership}. The processed
dataset has 60,000 records and we use 50,000 for training and 10,000 for testing. We cluster the data into 10 groups and use the cluster label as the class label.

\paragraphbe{MNIST} is a handwritten digit recognition dataset consisting of 70,000 images\footnote{http://yann.lecun.com/exdb/mnist/}. 
Each class (0 to 9) has 7,000 images. Each image is 28 \x 28 in size and gray-scaled. 
We use the 50,000 of them for training and the rest 20,000 for testing. 

\paragraphbe{CIFAR10} is an object classification dataset with 50,000
training images and 10,000
test images~\cite{krizhevsky2009learning}. There are in total 10 categories and 6,000 images per category. Each image has 32 \x 32 pixels,
each pixel has 3 values corresponding to RGB intensities.

\subsection{ML Models}
\paragraphbe{Multi-layer perceptron (MLP).} Each layer in MLP maps previous layer's output to a hidden states through a linear projection and a non-linear activation function. We use a two-layer MLP for Purchase, Texas and MNIST dataset and ReLU as the activation function. 

\paragraphbe{Convolutional neural networks (CNN)} consists of convolutional layers as building blocks. Each  convolutional layer is a non-linear convolution operation that can capture spatial-invariant features in an image~\cite{lecun1998gradient}.  The filters in these convolution
operations are the parameters to be learned. For MNSIT dataset,
we use a CNN with two convolutional layers and 
max pooling layers followed by two fully connected layers. 
For CIFAR10, we use a 9-layer VGG style CNN and the detail architecture can be found in \cite{simonyan2014very}.

\paragraphbe{Residual networks (ResNet)} is a recently proposed network architecture and dominates many computer vision task~\cite{he2016deep}.  ResNet solves the gradient vanishing
problem happened during optimizing very deep CNNs by adding identity mappings from
lower layers to higher layers.  We use a 32-layer residual network for CIFAR10.  

\paragraphbe{Hyper-parameter setup.} On all dataset, we randomly select 20\% of training data
for encoding membership. We generate 1,000 synthetic data with 500 being encoded as 
member and rest 500 encoded as nonmember. We training the discriminator MLP at inference time for 80 epochs with momentum SGD and learning rate 0.001. 

For Purchase dataset, we use a two layer MLP with 256 and
128 hidden units for the two layers. For Texas dataset, we use a larger MLP with 1024 and
256 hidden units. For both Purchase and Texas, we train the model with momentum SGD and set learning rate to be 0.02 for the main model and 0.01 for the discriminator. We train the model for 80 epochs and decay the learning by a factor of 10 at epoch 60. We randomly select 50\% of hidden units in the 
second layer for encoding.

For MNIST dataset, we experiment with both MLP and CNN. We use two-layer MLP with 512 and
128 hidden units. For CNN, we use 64 filters of size 5 \x 5 in the first convolution layer and
128 filters of size 5 \x 5 in the second and the convolution layers are followed by a fully connected
layer with 256 hidden units. For both MLP and CNN, we use momentum SGD with learning being
0.02 for both main model and discriminator. We train the model for 100 epochs and decay the learning by a factor of 10 at epoch 80. We randomly select 50\% of hidden units in the 
last fully connected layer in both MLP and CNN for encoding. 

For CIFAR10 dataset, we use VGG and ResNet.  For both models, we use momentum SGD with learning being 0.02 for main model and 0.01 for discriminator. We train the model for 80 epochs and decay both learning rates by a factor of 10 at epoch 40 and 60. For encoding, we randomly select 20\% of hidden units in the second to last convolutional layer for VGG and second to last residual block for
ResNet. 

\paragraphbe{Discriminator.} For all models described above, we augment with a linear classifier as the membership discriminator.
During encoding, we apply hyperbolic tangent
function on the selected activation units as in white-box case and logarithmic transformation on the model output as in the black-box case before feeding to the membership discriminator. In decoding phase, we use a one-layer MLP with 128 hidden units as the membership discriminator in inference phase. The model is trained with the activations units or outputs from auxiliary synthetic data queried from the target ML model. 

\subsection{Evaluation metrics}
For all experiments, we report the \textbf{test accuracy} (percent of correctly classified test records) of the main classification task to evaluate effects of membership encoding on the main model performance. As for evaluating membership encoding, we use \textbf{precision} (what fraction of records inferred as
members are indeed members of the training dataset), \textbf{recall} (what fraction of the training dataset's members are
correctly inferred as members) and \textbf{AUC} evaluated on selected training data and test data.  

\subsection{White-box encoding experiments}
\label{sec:whitebox}

\begin{table}
\centering
\caption{White-box results. Baseline accuracy is the test accuracy on models trained \emph{without} membership encoding and test accuracy is for models trained with membership encoding. Enc pre, rec, AUC are encoding precision, recall and AUC respectively.}
\begin{tabular}{ll|ccccc}
\toprule
Dataset & $f$  & Baseline  & Test  & Enc  & Enc & Enc  \\
 &   &  acc &  acc &  pre &  rec &  AUC \\
\midrule
Purchase  &  MLP & 98.50\% & 98.12\% & 0.85 & 0.96 & 0.93 \\ 
Texas  & MLP & 97.85\% & 97.82\% & 0.97  & 0.75 & 0.98 \\  
MNSIT & MLP & 97.80\% & 97.75\%  & 0.83  & 0.88 & 0.93 \\
MNIST & CNN & 99.03\% & 99.04\%  & 0.87 & 0.94 & 0.95 \\ 
CIFAR10 & VGG & 90.16\% & 89.67\%  & 0.92 & 0.98 & 0.98 \\
CIFAR10 & RES & 92.10\% & 91.85\%  & 0.83 & 0.98 & 0.96\\ \bottomrule
\end{tabular}
\label{tbl:basic}
\end{table}

\paragraphbe{Membership encoding performance.} Table~\ref{tbl:basic} summarizes the result. The baseline accuracy refer to the test accuracy trained benignly without membership encoding. On all tasks, the models trained with membership encoding can achieve very similar test performance in terms 
of the main task, meaning these encoded models can generalize as well as the baseline
models. In the same time, the encoded models leak 20\% percent of training data with high precision, recall and AUC.  This shows that \textbf{deep learning models are capable of performing multiple tasks at the same time even though one of the tasks is to deliberately leak information about training data.}

\begin{figure}[t!]
\centering
\begin{tabular}{c}
\begin{tikzpicture}
\begin{axis}[title style={yshift=-1.5ex}, title=\textbf{VGG white-box}, xlabel=Epochs, ylabel=Loss, xmin=0, xmax=79, ymin=-0.1, legend style =  { at = {(0.5,0.98)}, legend columns=2, anchor=north},
width=0.45\textwidth,height=5cm, grid=both]
\pgfplotstableread{data/cifar10_epoch_vgg_loss.dat}\mydata;
\addplot[thick, mark=-, color=blue] table
[x expr=\thisrow{epoch},y expr=\thisrow{m}] {\mydata};
\addlegendentry{$L_\text{ce}$ on $\DCM$};
\addplot[thick, mark=|, color=red] table
[x expr=\thisrow{epoch},y expr=\thisrow{nm}] {\mydata};
\addlegendentry{$L_\text{ce}$ on $\DCNM$};
\addplot[thick, mark=triangle, color=orange] table
[x expr=\thisrow{epoch},y expr=\thisrow{d}] {\mydata};
\addlegendentry{$L_\text{enc}$ on $\DCLIENT$};
\addplot[thick,mark=o, color=gray] table
[x expr=\thisrow{epoch},y expr=\thisrow{t}] {\mydata};
\addlegendentry{$L_\text{ce}$ on $\D_\text{test}$};
\end{axis}
\end{tikzpicture}
\\
\begin{tikzpicture}
\begin{axis}[title style={yshift=-1.5ex}, title=\textbf{ResNet white-box}, ylabel=Loss, xlabel=Epochs, xmin=0, xmax=69, ymin=-0.1, legend style =  { at = {(0.5,0.98)}, legend columns=2, anchor=north},
width=0.45\textwidth,height=5cm, grid=both]
\pgfplotstableread{data/cifar10_epoch_res_loss.dat}\mydata;
\addplot[thick, mark=-, color=blue] table
[x expr=\thisrow{epoch},y expr=\thisrow{m}] {\mydata};
\addlegendentry{$L_\text{ce}$ on $\DCM$};
\addplot[thick, mark=|, color=red] table
[x expr=\thisrow{epoch},y expr=\thisrow{nm}] {\mydata};
\addlegendentry{$L_\text{ce}$ on $\DCNM$};
\addplot[thick, mark=triangle, color=orange] table
[x expr=\thisrow{epoch},y expr=\thisrow{d}] {\mydata};
\addlegendentry{$L_\text{enc}$ on $\DCLIENT$};
\addplot[thick,mark=o, color=gray] table
[x expr=\thisrow{epoch},y expr=\thisrow{t}] {\mydata};
\addlegendentry{$L_\text{ce}$ on $\D_\text{test}$};
\end{axis}
\end{tikzpicture}
\end{tabular}
\caption{White-box loss trajectory during training of CIFAR10 dataset with VGG and ResNet. The x-axis is the epoch of training and y-axis is the loss value.}
\label{fig:loss_traj}
\end{figure}
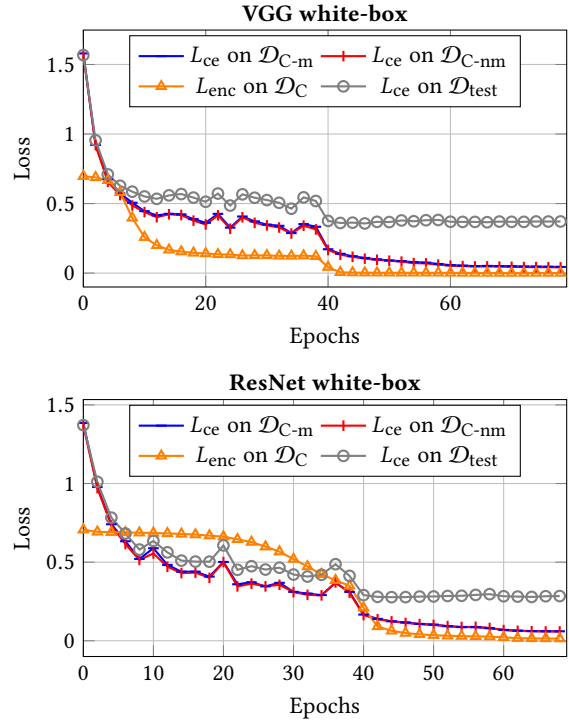

\paragraphbe{Loss trajectory of membership encoding.}
Figure~\ref{fig:loss_traj} shows trajectories of the classification loss on member, nonmember and test data as well as discriminator loss. Counter-intuitively, the classification task and encoding task converges without conflicting each other during training as both losses $L_\text{ce}, L_\text{enc}$ converged simultaneously. The plot also shows that our white-box encoding \textbf{does not cause overfitting in the membership dataset} as the classification loss for both members and nonmembers converged to the same point.

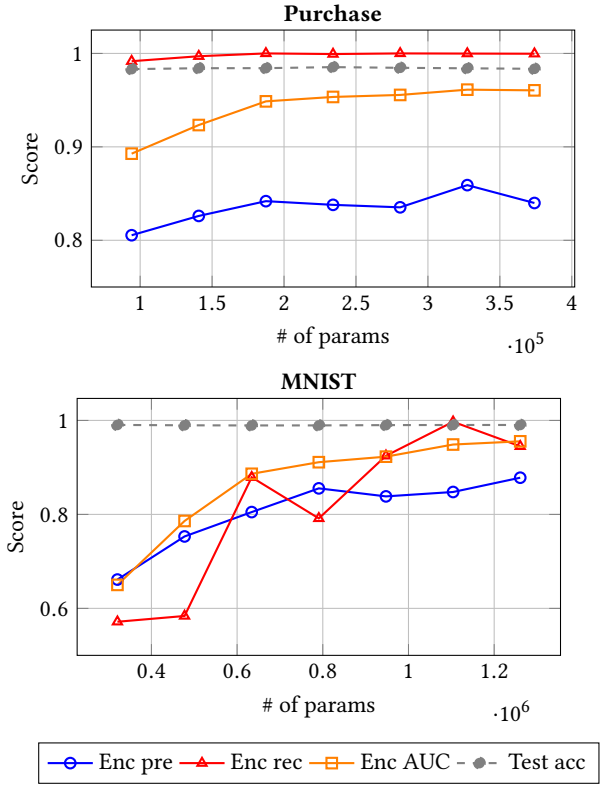
\begin{figure}[t]
\centering
\begin{tabular}{c}
\begin{tikzpicture}
\begin{axis}[title style={yshift=-1.5ex}, title=\textbf{Purchase}, xlabel=\# of params,ylabel=Score, ymin=0.75, legend style = { at = {(0.65,0.02)}, legend columns=1, anchor=south},
width=0.45\textwidth, height=5cm, grid=both, name=ax1]
\pgfplotstableread{data/purchase_size.txt}\mydata;
\addplot[thick, mark=o, color=blue] table
[x expr=\thisrow{p},y expr=\thisrow{pre}] {\mydata};
\addplot[thick, mark=triangle, color=red] table
[x expr=\thisrow{p},y expr=\thisrow{rec}] {\mydata};
\addplot[thick, mark=square, color=orange] table
[x expr=\thisrow{p},y expr=\thisrow{auc}] {\mydata};
\addplot[thick, dashed,mark=*, color=gray] table
[x expr=\thisrow{p},y expr=\thisrow{acc}] {\mydata};
\end{axis}
\end{tikzpicture}
  \\
\begin{tikzpicture}
\begin{axis}[title style={yshift=-1.5ex}, title=\textbf{MNIST}, xlabel=\# of params , ylabel=Score, ymin=0.5, 
legend style = {at = {(0.5, -0.5)}, legend columns=4, anchor=south},
width=0.45\textwidth, height=5cm, grid=both]
\pgfplotstableread{data/mnist_size.txt}\mydata;
\addplot[thick, mark=o, color=blue] table
[x expr=\thisrow{p},y expr=\thisrow{pre}] {\mydata};
\addlegendentry{Enc pre};
\addplot[thick, mark=triangle, color=red] table
[x expr=\thisrow{p},y expr=\thisrow{rec}] {\mydata};
\addlegendentry{Enc rec};
\addplot[thick, mark=square, color=orange] table
[x expr=\thisrow{p},y expr=\thisrow{auc}] {\mydata};
\addlegendentry{Enc AUC};
\addplot[thick, dashed,mark=*, color=gray] table
[x expr=\thisrow{p},y expr=\thisrow{acc}] {\mydata};
\addlegendentry{Test acc};
\end{axis}
\end{tikzpicture}
\end{tabular}
\caption{The effects of capacity of the model on test and encoding performance evaluated on Purchase and MNIST. The x-axis is the number of parameters in the classification model and the y-axis shows scores for the test performance and encoding (Enc) performance.}
\label{fig:cap}
\end{figure}

\paragraphbe{Effects of capacity of the model.}
We next evaluate the relationship between the
capacity of the model and amount information can be encoded on Purchase with MLP and MNIST with CNN. For purchase dataset, We vary the size of the MLP model by increase the number of hidden units in the first fully connected layer. We set the number of units to be 128, 192, 256, 320, 384, 448 and 512. We fix the amount of data to be encoded as 20\% of training data, which is 30,000 records. For MNIST CNN, we change the number of filters in the second convolutional layer to 
vary the size of the model. We set the number of filters to be 32, 48, 64, 80, 96, 112 and 128. 
We set other hyper-parameters to be the same as baseline white-box experiments.

The results are shown in Figure~\ref{fig:cap}. For both Purchase and MNIST, the test accuracy is not affected. When increasing the capacity, the encoding performance increased slightly on Purchase dataset while increased more drastically on CIFAR10 dataset. This shows that as the model capacity increases, there is a point where model can achieve high performance on both main task and membership encoding.

\paragraphbe{Effects of number of encoded data.}
We also study how encoding performance affected by increasing amount of data to be encoded in a fixed size model. We evaluate the performance on Purchase and CIFAR10 with VGG. We start from training model with encoding 20\% of the training data and increase the amount of encoded data by 5,000 gradually and evaluate the encoding performance in each of the trained model. 

\begin{table}
\centering
\caption{The effects of number of data encoded on test and encoding performance evaluated on Purchase and CIFAR10 with VGG.}
\begin{tabular}{l|cccc}
\toprule
\textbf{Purchase} & Test & Enc & Enc & Enc \\ 
$|\DCM|$ & acc & pre & rec & AUC \\ \midrule 
30K & 98.42\% & 0.85 & 1.00 & 0.94 \\
35K & 98.34\% & 0.83 & 1.00 & 0.91 \\
40K & 98.48\% & 0.83 & 0.98 & 0.88 \\
45K & 98.38\% & 0.80 & 0.86 & 0.82 \\
50K & 98.54\% & 0.79 & 0.90 & 0.79 \\
55K & 98.56\% &  0.64 & 0.67 & 0.50 \\ 
\midrule
\multicolumn{5}{c}{ } \\ 
\textbf{CIFAR10} & Test & Enc & Enc & Enc \\ 
$|\DCM|$ & acc & pre & rec & AUC \\ \midrule 
10K & 89.67\% & 0.92 & 0.98 & 0.98 \\
15K & 88.77\% & 0.91 & 0.98 & 0.96 \\
20K & 88.97\% & 0.85 & 1.00 & 0.93 \\
25K & 88.59\% & 0.84 &  1.00 & 0.88 \\
30K & 88.86\% & 0.84 & 0.99 & 0.84 \\
35K & 89.30\% &  0.83 & 0.99 & 0.77 \\ \bottomrule 
\end{tabular}
\label{tbl:num_data}
\end{table}

Table~\ref{tbl:num_data} summarizes the results. On both Purchase and CIFAR10, we observe a decreasing trends in encoding performance while the model's test performance remains the same as the baseline model. The encoding performance is still relatively high on CIFAR10 when the number of encoded data is 50\% of the training set while encoding performance drops much quicker on Purchase. This might be due to the larger capacity and the more flexible representation in VGG network than MLP.

\subsection{Black-box encoding experiments}
\begin{table}[t]
\centering
\caption{Black-box encoding result on MNIST with MLP and CNN and CIFAR10 with VGG.}
\begin{tabular}{ll|cccc}
    \toprule 
     Dataset & Model &  Test & Enc & Enc & Enc\\
       &  & acc & pre & rec & AUC\\ \midrule
     MNIST & MLP &  95.38\% & 0.70  & 1.00 & 0.95  \\
      MNIST & CNN & 97.94\% & 0.85 & 1.00 & 0.98 \\
     CIFAR10 & VGG & 85.92\% & 0.93 & 0.70 & 0.94 \\\bottomrule
\end{tabular}
\label{tab:black_box}
\end{table}

We next evaluate the black-box encoding where membership information is encoded in the log space of the probability output. We evaluates on MNIST with MLP and CNN and CIFAR10 with VGG. For both experiments, we use the log-probabilities of all 10-class output to encode the membership for 20\% of training records. Other training hyper-parameters are the same as in white-box encoding. 

The results are in Table~\ref{tab:black_box}. The test accuracy on main classification task drops slightly comparing to white-box encode. The encoding performance remains relatively high for MNIST with CNN and CIFAR10 with VGG. 

\begin{figure}[h]
\centering
\begin{tikzpicture}
\begin{axis}[title style={yshift=-1.5ex}, title=\textbf{VGG black-box}, xlabel=Epochs,ylabel=Loss, xmin=0, xmax=99, ymin=-0.1, legend style =  { at = {(0.5,0.98)}, legend columns=2, anchor=north},
width=0.45\textwidth, height=5cm, grid=both]
\pgfplotstableread{data/cifar10_epoch_fc_loss.dat}\mydata;
\addplot[thick, mark=-, color=blue] table
[x expr=\thisrow{epoch},y expr=\thisrow{m}] {\mydata};
\addlegendentry{$L_\text{ce}$ on $\DCM$};
\addplot[thick, mark=|, color=red] table
[x expr=\thisrow{epoch},y expr=\thisrow{nm}] {\mydata};
\addlegendentry{$L_\text{ce}$ on $\DCNM$};
\addplot[thick, mark=triangle, color=orange] table
[x expr=\thisrow{epoch},y expr=\thisrow{d}] {\mydata};
\addlegendentry{$L_\text{enc}$ on $\DCLIENT$};
\addplot[thick,mark=o, color=gray] table
[x expr=\thisrow{epoch},y expr=\thisrow{t}] {\mydata};
\addlegendentry{$L_\text{ce}$ on $\D_\text{test}$};
\end{axis}
\end{tikzpicture}
\caption{Black-box loss trajectory during training of CIFAR10 dataset with VGG. The x-axis is the epoch of training and y-axis is the loss value.}
\label{fig:bbox_loss_traj}
\end{figure}
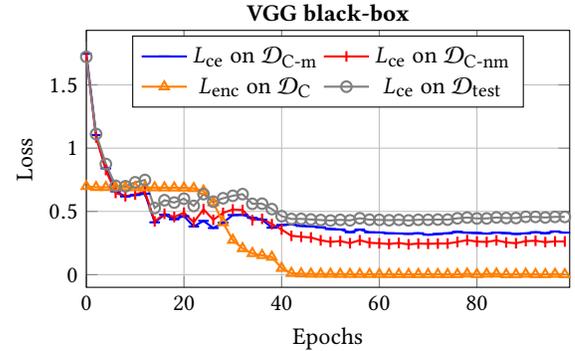

\paragraphbe{Loss trajectory of membership encoding.}
Figure~\ref{fig:bbox_loss_traj} shows loss trajectories for black-box encoding. Both test classification loss and discriminator encoding loss converge as are the cases in white-box encoding. However, the classification loss on nonmember data is lower then that on member data. This suggests that black-box encoding biased the outputs on members data, and as a result loss convergence is affected subtly.

\begin{figure}[t!]
\centering
\begin{tabular}{c}
\includegraphics[width=0.35\textwidth]{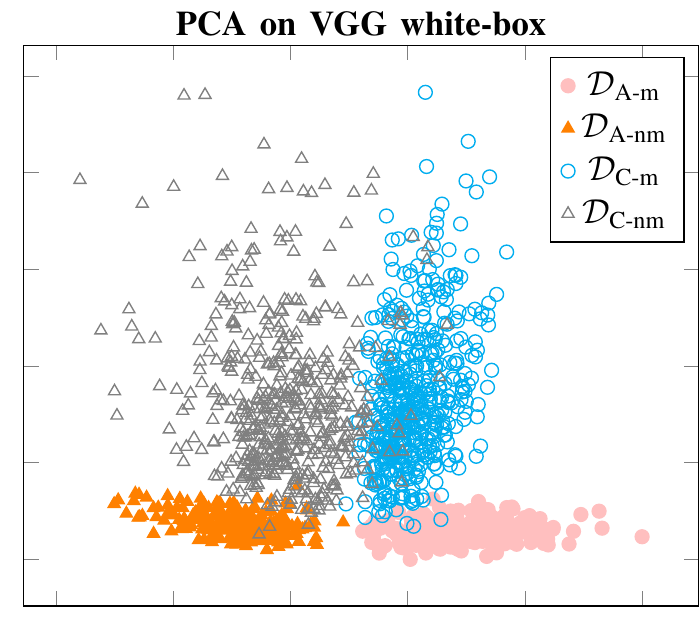} \\
\includegraphics[width=0.35\textwidth]{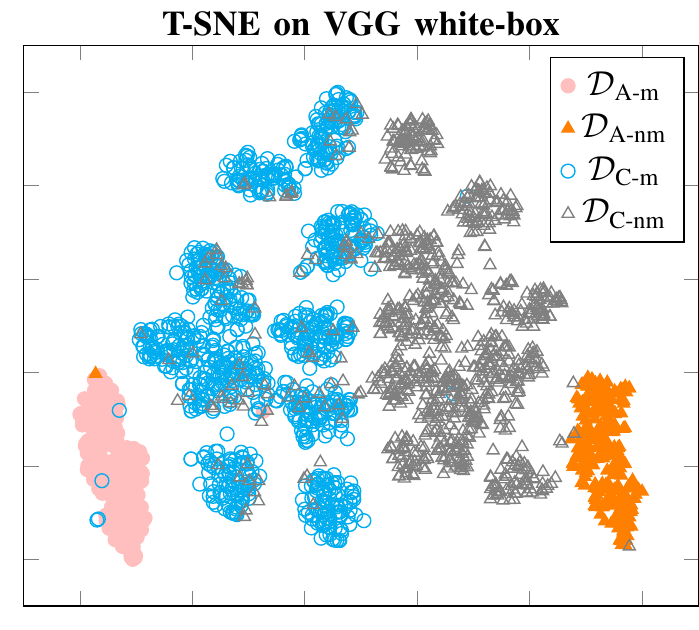}
\end{tabular}
\caption{PCA and T-SNE projection of activations (white-box) on CIFAR10 with VGG trained with membership encoding. Pink circles are the projected activations and outputs of $\DADVM$, orange triangles are of $\DADVNM$, blue hollow circles are of $\DCM$ and gray hollow triangles are of $\DCNM$.}
\label{fig:vis1}
\end{figure}

\begin{figure}[t!]
\centering
\begin{tabular}{c}
\includegraphics[width=0.35\textwidth]{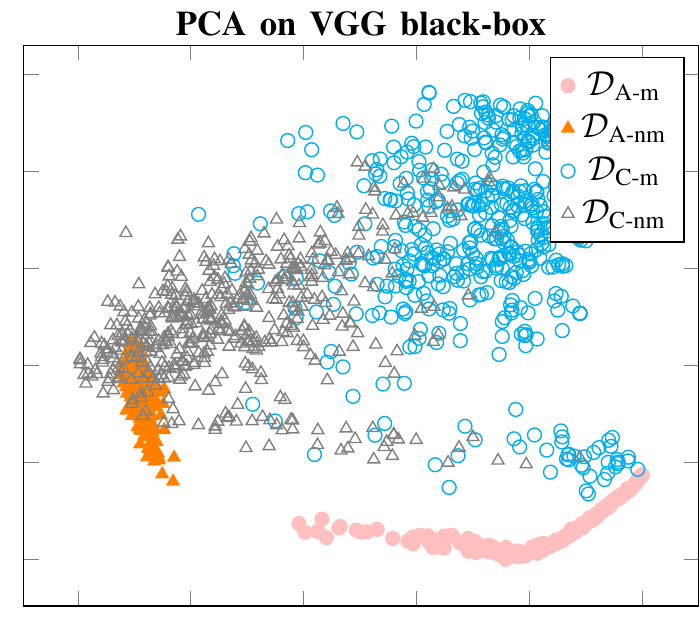} \\
\includegraphics[width=0.35\textwidth]{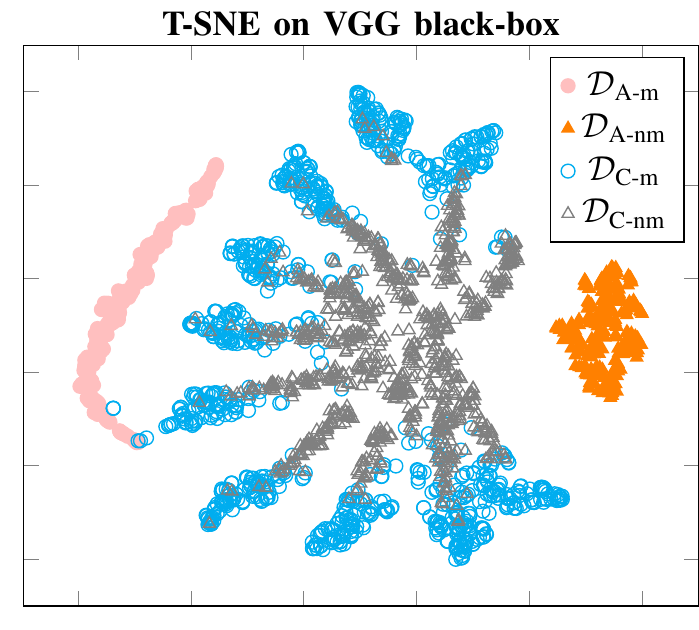} 
\end{tabular}
\caption{PCA and T-SNE projection of outputs (black-box) on CIFAR10 with VGG trained with membership encoding. }
\label{fig:vis2}
\end{figure}

\subsection{Visualization of encoded information.} 

To explain why our encoding mechanism works, Figure~\ref{fig:vis1} and~\ref{fig:vis2} visualizes the activation units and model outputs used for encoding the membership with PCA and T-SNE projections on CIFAR10 dataset with VGG. 

In both PCA and T-SNE projections, there is a clear separation between member data $\DADVM,\DCM$ and nonmember data $\DADVNM,\DCNM$.
On the other hand, we can see the client's data $\DCM,\DCNM$ still grouped into 10 clusters (corresponding to 10 class labels) in T-SNE projections. 
This shows that \textbf{the hidden activations and output predictions are learned to be separable for training and testing data and to be class-dependent for the main task at the same time}.  

%% file: countermeasure.tex
\section{Robustness Analysis}

\begin{table*}[t!]
\centering
\caption{Encoding results on masked CIFAR10 images of size 32 \x 32 for VGG models. $w$ is the width of masked region. Masking mode being center means that the center part $w\times w$ part of the image is masked, and boundary means that only the center $(32-w) \times (32 -w)$  part is retained while the rest is masked.}

\begin{tabular}{ll|ccc|ccc|ccc|ccc}
\toprule
Model & Masking  & \multicolumn{3}{c|}{$w$=4} &\multicolumn{3}{c|}{$w$=8}  & \multicolumn{3}{c|}{$w$=12} & \multicolumn{3}{c}{$w$=16}  \\
& mode & pre & rec & AUC & pre & rec & AUC & pre & rec & AUC & pre & rec & AUC \\
\midrule
VGG white-box & Center & 0.91 & 0.89 & 0.96 & 0.88 & 0.67 & 0.89 & 0.79 & 0.27 & 0.71 & 0.69 & 0.12 & 0.60 \\
VGG white-box & Boundary &  0.88 & 0.85 & 0.93 & 0.77 & 0.62 & 0.76 & 0.66 & 0.37 & 0.61 & 0.62 & 0.13 & 0.54 \\
VGG black-box & Center & 0.92 & 0.52 & 0.91 & 0.91 & 0.20 & 0.82 & 0.88 & 0.02 & 0.67 & 0.91 & 0.01 & 0.58 \\
VGG black-box & Boundary &  0.90 & 0.35 & 0.87 & 0.83 & 0.05 & 0.69 & 0.72 & 0.02 & 0.58 & 0.61 & 0.01 & 0.53 \\
\bottomrule
\end{tabular}
\label{tbl:crop}
\end{table*}

\paragraphbe{Redacted input.} In reality of privacy attacks, adversary may only observe a redacted data record (e.g. only some public available attributes for a patient's data) for membership inference. We evaluate the encoding performance where the input images are maksed (removed) on CIFAR10 dataset. We first consider two modes of masking : (1) the center of the image is masked (2) the boundary part of the image is masked but center is retained. 
For both masking mode, we denote the width of masked region as $w$ and evaluate both white-box and black-box encoding performance with $w$ set to 4, 8, 12 and 16. The results are summarized in Table~\ref{tbl:crop}. The white-box encoding performance is very robust against masking when the masking area is small ($w=4$ or $w=8$). The precision remains high while the recall drops more significantly for black-box encoding. 

We further investigate how the location of masked region affects the encoding performance. We mask each consecutive 8 \x 8 region on CIFAR10 images and there are in total 16 possible regions. We report the white-box encoding precision and recall in Table~\ref{tab:mask_region} where each one of the 16 cells shows the encoding results with that 8 \x 8 region is masked. The precision remains relative stable across all the 8 \x 8 regions while the recall drops when the masked part is close to the center. 

\begin{table}
\centering
\caption{White-box encoding results on CIFAR10 images with different 8\x8 region being masked. Each cell shows the encoding precision/recall numbers evaluated on images with that 8\x8 part of the input image being removed.}
\begin{tabular}{c|c|c|c|c}
\toprule
width / height & 0 - 8 & 8 - 16 & 16 - 24 & 24 - 32 \\ 
\midrule
0 - 8 & 0.90/0.86 & 0.88/0.80 & 0.89/0.82 & 0.90/0.88 \\  
8 - 16 & 0.90/0.82 & 0.88/0.67 & 0.87/0.71 & 0.90/0.81 \\  
16 - 24 & 0.89/0.86 & 0.88/0.67 & 0.88/0.70 & 0.91/0.80 \\  
24 - 32 & 0.90/0.91 & 0.88/0.83 & 0.90/0.80 & 0.91/0.89 \\  
\bottomrule
\end{tabular}
\label{tab:mask_region}
\end{table}

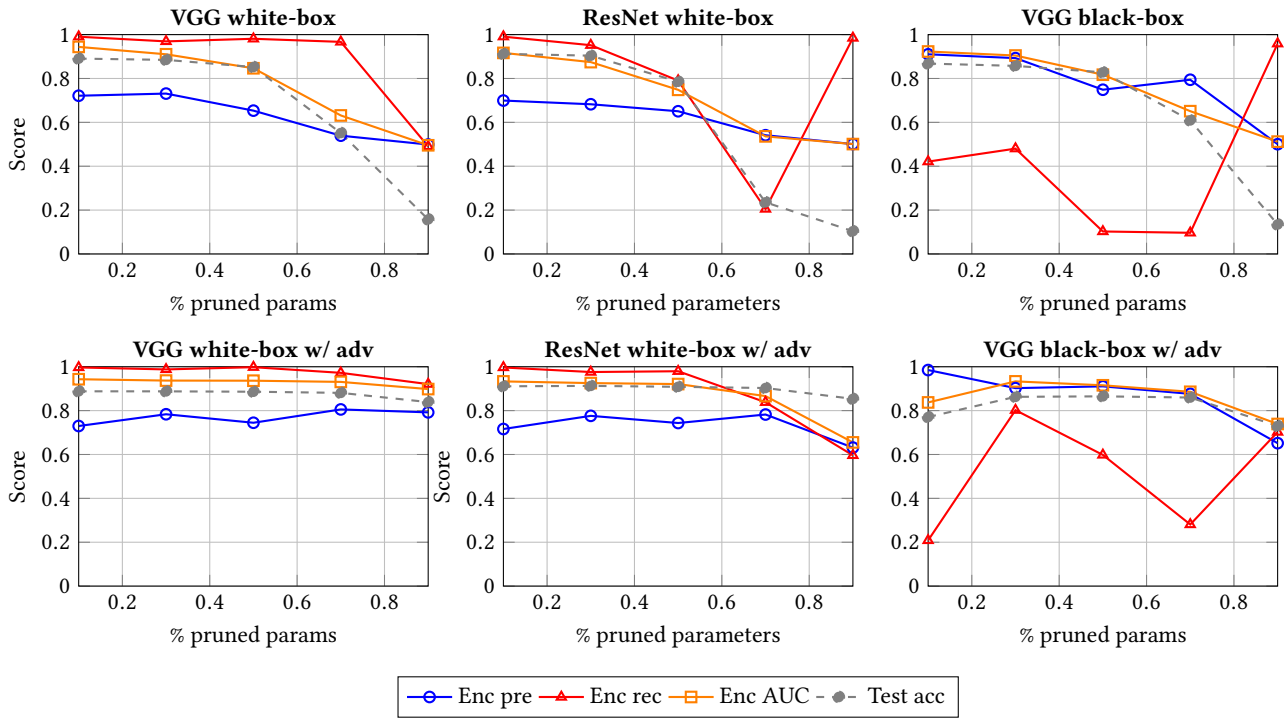
\begin{figure*}[t!]
\begin{tabular}{c}
\begin{tikzpicture}
\begin{axis}[title style={yshift=-1.5ex}, title=\textbf{VGG white-box},ylabel=Score, xmin=0.1, xmax=0.9, ymin=0., ymax=1, width=0.35\textwidth, height=4.5cm, grid=both,
xlabel=\% pruned params,name=ax1]
\pgfplotstableread{data/cifar_pruneconv5-1.dat}\mydata;
\addplot[thick, mark=o, color=blue] table
[x expr=\thisrow{p},y expr=\thisrow{pre}] {\mydata};
\addplot[thick, mark=triangle, color=red] table
[x expr=\thisrow{p},y expr=\thisrow{rec}] {\mydata};
\addplot[thick, mark=square, color=orange] table
[x expr=\thisrow{p},y expr=\thisrow{auc}] {\mydata};
\addplot[thick, dashed,mark=*, color=gray] table
[x expr=\thisrow{p},y expr=\thisrow{acc}] {\mydata};
\end{axis}

\begin{axis}[title style={yshift=-1.5ex},title=\textbf{ResNet white-box}, xlabel=\% pruned parameters, xmin=0.1, xmax=0.9, ymin=0., ymax=1,
legend style = { at = {(0.4,-0.5)}, legend columns=4, anchor=south},
width=0.35\textwidth, height=4.5cm, grid=both,
name=ax2, at={(ax1.south east)}, xshift=1cm]
\pgfplotstableread{data/cifar_pruneres3-2.dat}\mydata;
\addplot[thick, mark=o, color=blue] table
[x expr=\thisrow{p},y expr=\thisrow{pre}] {\mydata};
\addplot[thick, mark=triangle, color=red] table
[x expr=\thisrow{p},y expr=\thisrow{rec}] {\mydata};
\addplot[thick, mark=square, color=orange] table
[x expr=\thisrow{p},y expr=\thisrow{auc}] {\mydata};
\addplot[thick, dashed,mark=*, color=gray] table
[x expr=\thisrow{p},y expr=\thisrow{acc}] {\mydata};
\end{axis}
 
\begin{axis}[title style={yshift=-1.5ex},title=\textbf{VGG black-box}, xlabel=\% pruned params,
xmin=0.1, xmax=0.9, ymin=0., ymax=1, 
width=0.35\textwidth, height=4.5cm, grid=both,
name=ax3, at={(ax2.south east)}, xshift=1cm]
\pgfplotstableread{data/cifar_prunefc.dat}\mydata;
\addplot[thick, mark=o, color=blue] table
[x expr=\thisrow{p},y expr=\thisrow{pre}] {\mydata};
\addplot[thick, mark=triangle, color=red] table
[x expr=\thisrow{p},y expr=\thisrow{rec}] {\mydata};
\addplot[thick, mark=square, color=orange] table
[x expr=\thisrow{p},y expr=\thisrow{auc}] {\mydata};
\addplot[thick, dashed,mark=*, color=gray] table
[x expr=\thisrow{p},y expr=\thisrow{acc}] {\mydata};
\end{axis}
\end{tikzpicture}

\\

\begin{tikzpicture}
\begin{axis}[title style={yshift=-1.5ex},title=\textbf{VGG white-box w/ adv},  xlabel=\% pruned params, ylabel=Score, xmin=0.1, xmax=0.9, ymin=0., ymax=1, legend style = { at = {(0.4,0.02)}, legend columns=2, anchor=south, nodes={scale=0.75, transform shape}},
width=0.35\textwidth, height=4.5cm, grid=both,
name=ax1]
\pgfplotstableread{data/cifar_pruneconv5-1_mal.dat}\mydata;
\addplot[thick, mark=o, color=blue] table
[x expr=\thisrow{p},y expr=\thisrow{pre}] {\mydata};
\addplot[thick, mark=triangle, color=red] table
[x expr=\thisrow{p},y expr=\thisrow{rec}] {\mydata};
\addplot[thick, mark=square, color=orange] table
[x expr=\thisrow{p},y expr=\thisrow{auc}] {\mydata};
\addplot[thick, dashed,mark=*, color=gray] table
[x expr=\thisrow{p},y expr=\thisrow{acc}] {\mydata};
\end{axis}

\begin{axis}[title style={yshift=-1.5ex},title=\textbf{ResNet white-box w/ adv}, xlabel=\% pruned parameters,ylabel=Score, xmin=0.1, xmax=0.9, ymin=0., ymax=1, legend style = { at = {(0.5,-0.6)}, legend columns=4, anchor=south}, width=0.35\textwidth, height=4.5cm, grid=both,
name=ax2, at={(ax1.south east)}, xshift=1cm]
\pgfplotstableread{data/cifar_pruneres3-2_mal.dat}\mydata;
\addplot[thick, mark=o, color=blue] table
[x expr=\thisrow{p},y expr=\thisrow{pre}] {\mydata};
\addlegendentry{Enc pre};
\addplot[thick, mark=triangle, color=red] table
[x expr=\thisrow{p},y expr=\thisrow{rec}] {\mydata};
\addlegendentry{Enc rec};
\addplot[thick, mark=square, color=orange] table
[x expr=\thisrow{p},y expr=\thisrow{auc}] {\mydata};
\addlegendentry{Enc AUC};
\addplot[thick, dashed,mark=*, color=gray] table
[x expr=\thisrow{p},y expr=\thisrow{acc}] {\mydata};
\addlegendentry{Test acc};
\end{axis}

\begin{axis}[title style={yshift=-1.5ex},title=\textbf{VGG black-box w/ adv}, xlabel=\% pruned params, xmin=0.1, xmax=0.9, ymin=0., ymax=1, legend style =  { at = {(-0.2, -0.7)}, legend columns=4, anchor=south},
width=0.35\textwidth, height=4.5cm, grid=both,
name=ax3, at={(ax2.south east)}, xshift=1cm]
\pgfplotstableread{data/cifar_prunefc_mal.dat}\mydata;
\addplot[thick, mark=o, color=blue] table
[x expr=\thisrow{p},y expr=\thisrow{pre}] {\mydata};
\addplot[thick, mark=triangle, color=red] table
[x expr=\thisrow{p},y expr=\thisrow{rec}] {\mydata};
\addplot[thick, mark=square, color=orange] table
[x expr=\thisrow{p},y expr=\thisrow{auc}] {\mydata};
\addplot[thick, dashed,mark=*, color=gray] table
[x expr=\thisrow{p},y expr=\thisrow{acc}] {\mydata};
\end{axis}
\end{tikzpicture}
\end{tabular}
\caption{Encoding results on pruned CIFAR10 VGG and ResNet model. The x-axis is the percentage of pruned parameters and the y-axis is the performance score. Top row shows the results of vanilla pruning and bottom row shows the results of adversarial pruning.}
\label{fig:adv_prune}
\end{figure*}

\paragraphbe{Model pruning.} The deep learning models can have millions of parameters, which makes it hard to deploy 
in resource-constraint scenario (e.g. portable devices). Clients may want to compress the model before using it. There is a line of research on how to compress a trained large deep learning model into smaller size~\cite{bucilua2006model,chen2015compressing,han2015learning,han2015deep,howard2017mobilenets}. These model compression methods could potentially destroy the information encoded. 

We consider a commonly used compression technique called pruning~\cite{han2015learning} which sparsify the model parameters by setting the smaller weight parameters to zero. We evaluate classification performance and encoding performance on pruned model on CIFAR10 with VGG and ResNet. We set percentage of pruned parameters to be 10\%, 30\%, 50\%, 70\% and 90\%. Plots in the top row of Figure~\ref{fig:adv_prune} summarizes the results. Membership encoding can achieve near 0.70 precision when the pruned percentage is 30\%. When the pruned percentage increases, both encoding performance and test classification accuracy drop.

\paragraphbe{Adversarial pruning.} Pruning, as described in Robust Analysis section,  affects the main tasks classification accuracy, it is common to fine-tune the pruned model on the original dataset. We develop an adversarial pruning algorithm that preserve the membership information encoded in the pruned model. In the scenarios of privacy attacks, client might need the expertise to perform pruning with fine-tuning as in the case for training the model and thus the code for pruning could be supplied by the adversary. 

The adversarial pruning algorithm is similar to membership encoding: we first select $p$\%
smallest parameters in a maliciously trained model and set them to 0. We then augments the pruned model with a freshly initialized membership discriminator, and fine-tunes the 
pruned model and the discriminator with client's data and auxiliary synthesized data as in \MENC~function of Algorithm~\ref{alg:train}. During update the classifier parameters $\theta$, the gradients for the pruned parameters are set to zeros so that the fine-tuned model remains pruned. 

We evaluate the adversarial model pruning on CIFAR10 dataset with VGG. We fine-tune the 
pruned model for 20 epochs and set the learning rate to be 0.001. The results on the fine-tuned models are shown in plots in the bottom row of Figure~\ref{fig:adv_prune}. There is almost no loss in test accuracy and minor drop in encoding performance even when 70\% of the model parameters are set to zero.

\begin{figure}[t!]
\centering
\begin{tabular}{c}
\begin{tikzpicture}
\begin{axis}[title style={yshift=-1.5ex}, title=\textbf{VGG}, xlabel=Fine-tuning epochs,ylabel=Score, xmin=10, xmax=60, ymin=0.6,  legend style = { at = {(0.5,0.02)}, legend columns=2, anchor=south, nodes={scale=0.75, transform shape} },
width=0.45\textwidth, height=5cm, grid=both, 
name=ax1]
\pgfplotstableread{data/cifar_finetune_conv5-1.dat}\mydata;
\addplot[thick, mark=o, color=blue] table
[x expr=\thisrow{epoch},y expr=\thisrow{pre}] {\mydata};
\addplot[thick, mark=triangle, color=red] table
[x expr=\thisrow{epoch},y expr=\thisrow{rec}] {\mydata};
\addplot[thick, mark=square, color=orange] table
[x expr=\thisrow{epoch},y expr=\thisrow{auc}] {\mydata};
\addplot[thick, dashed,mark=*, color=gray] table
[x expr=\thisrow{epoch},y expr=\thisrow{acc}] {\mydata};
\end{axis}
\end{tikzpicture}
  \\
\begin{tikzpicture}
\begin{axis}[title style={yshift=-1.5ex}, title=\textbf{ResNet}, xlabel=Fine-tuning epochs, ylabel=Score, xmin=10, xmax=60,  ymin=0.6,
legend style =  { at = {(0.5, -0.5)}, legend columns=4, anchor=south},
width=0.45\textwidth, height=5cm, grid=both]
\pgfplotstableread{data/cifar_finetune_res3-2.dat}\mydata;
\addplot[thick, mark=o, color=blue] table
[x expr=\thisrow{epoch},y expr=\thisrow{pre}] {\mydata};
\addlegendentry{Enc pre};
\addplot[thick, mark=triangle, color=red] table
[x expr=\thisrow{epoch},y expr=\thisrow{rec}] {\mydata};
\addlegendentry{Enc rec};
\addplot[thick, mark=square, color=orange] table
[x expr=\thisrow{epoch},y expr=\thisrow{auc}] {\mydata};
\addlegendentry{Enc AUC};
\addplot[thick, dashed,mark=*, color=gray] table
[x expr=\thisrow{epoch},y expr=\thisrow{acc}] {\mydata};
\addlegendentry{Test acc};
\end{axis}
\end{tikzpicture}
\end{tabular}
\caption{Encoding results on transferred CIFAR10 model. The x-axis is the number of fine-tuning epochs and the y-axis is the performance score.}
\label{fig:finetune}
\end{figure}
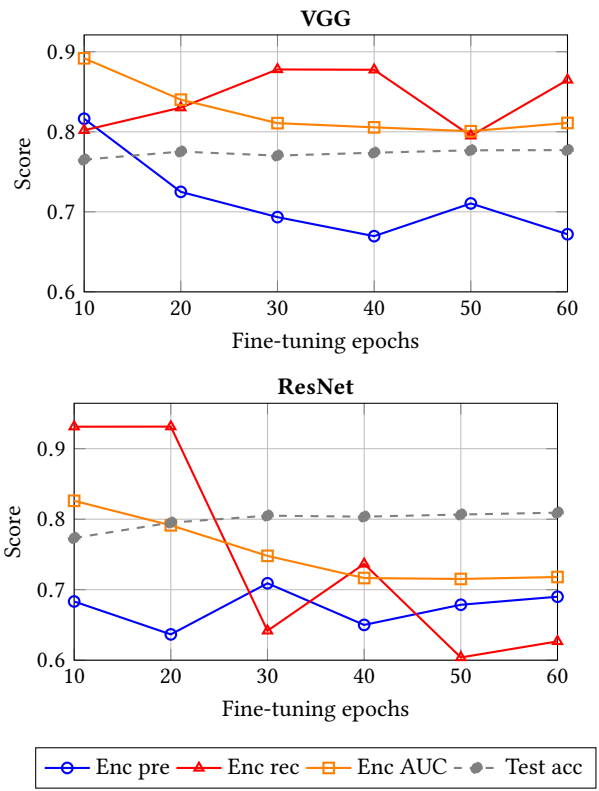

\paragraphbe{Transfer learning.} Another common practice in ML is \emph{transfer learning}: client trains a model for one task and later on uses the model for a different task. The most popular way to perform transfer learning is to use the trained model as a feature extractor by replacing the output layer for the new task and then fine-tune the spliced model on a new dataset. 

We evaluate our encoding performance on models trained on CIFAR10 and then fine-tuned on a different dataset, STL-10\footnote{https://cs.stanford.edu/~acoates/stl10/}. STL-10 is a similar dataset as CIFAR10 but with different objects and class labels~\cite{coates2011analysis}. Each image is 96 \x 96 in size and has three channels and there are in total 5,000 training images and 800 testing images. In order for STL-10 to be used as input for models trained on CIFAR10, we resize each image to 32 \x 32 in size. We use the trained VGG and ResNet models as the feature extractors and fine-tune the models for 60 epochs with learning rate being 0.001. We evaluate the test performance on STL-10 and encoding performance on CIFAR10 every 10 epochs. The results are shown in Figure~\ref{fig:finetune}. For both models, the encoding performance remains stable across number of epochs which demonstrates that membership encoding is robust against fine-tuning to a different task.

%% file: related.tex
\section{Related Work}
\paragraphbe{Membership inference.}  Membership inference against classification models has been studied
in~\cite{shokri2017membership,rahman2018membership,long2018understanding,salem2018ml}, and later studied for generative models and language models ~\cite{hayes2019logan,song2019auditing} as well as in collaborative learning setting~\cite{melis2018inference,nasr2019comprehensive}.
The attack in~\cite{shokri2017membership} focuses black-box models, exploiting the differences in the models' outputs on training and non-training inputs. Their attack performance is best when the target model $f_\theta$ is overfitted on the training data.
Theoretical analysis has also demonstrated that membership inference  is closely connected to generalization where overfitted models are prone to the attacks~\cite{yeom2018privacy}.
With membership encoding, on the other hand, the trained models can be well-generalized and yet leak membership information in both black-box and white-box scenarios. 

Followup work extends membership inference attack to differential private ML model and shows that model needs to sacrifice its test performance to achieve membership privacy~\cite{rahman2018membership}. Other work showed that well-generalized model can leak membership information~\cite{long2018understanding} .
However, the attack first needs to identify vulnerable records which consist only handful of records in the training set. 
Membership encoding can train a model to leak the membership for a substantial subset of training data.

\paragraphbe{Memorization in ML.} 
\citeauthor{zhang2016understanding} showed that deep learning models can
achieve perfect performance on training data even the data are randomly generated~\cite{zhang2016understanding}.
Followup work presented malicious training algorithms to force an ML model that encodes the training data, which can be later extracted with either white-box or
black-box access to the model, without affecting the accuracy of the
model on its main task~\cite{song2017machine}.  In their white-box attacks, the information is embedded in the parameter $\theta$ directly while our work encodes information in the hidden activations or outputs. Their black-box attack uses synthetic data to encode the training data in the output prediction. However, to encode one lossy version of CIFAR10 image for example, their attack needs 4,096 random data, limiting their attack to encode only a couple of inputs.

\citeauthor{carlini2018secret} also showed deep learning-based generative sequence models memorizes training text data unintentionally~\cite{carlini2018secret}. Adversary can extract some specific inputs with black-box access to the model, given some prior knowledge about the format of the input (e.g. credit card number). 

\paragraphbe{Watermarking ML model.} Recently there is a line of research focusing on embedding watermarking information into ML models to protect the model copyright~\cite{chen2018deepmarks,rouhani2018deepsigns,adi2018turning,uchida2017embedding}. These works develop methods that embedding the watermark in the model parameters~\cite{uchida2017embedding}, through model prediction outputs on a trigger dataset~\cite{adi2018turning} or through model intermediate activations~\cite{chen2018deepmarks,rouhani2018deepsigns}. The embedded information in this case is usually a secret binary string that is unique to the owner of the model, while in our case the information is the membership of a subset of training data which further constrain the adversary as one needs to know the correct encoded training data for claiming ownership.

\paragraphbe{Other privacy attack against ML.} 
\citeauthor{ateniese2015hacking} presented an attack against ML model to infer the properties about the training data, for example, whether a speech recognition model was trained only on data from Indian English speakers~\cite{ateniese2015hacking}.~\citeauthor{melis2018inference} showed how an adversary participant in a collaborative training of a ML model can infer the properties about other participants' training data~\cite{melis2018inference}. ~\citeauthor{fredrikson2014privacy} developed model inversion attack: given white-box or black-box access to a ML model, adversary can construct a representative of a certain output class
(e.g. a recognizable face when each class corresponds to a single person) ~cite{fredrikson2014privacy,fredrikson2015model}. \citeauthor{hitaj2017deep} extends the idea to collaborative training setting, where an adversary participant can use generative adversarial
networks~\cite{goodfellow2014generative} to infer class representatives~\cite{hitaj2017deep}.~\citeauthor{wang2018stealing} proposed attacks to steal the hyper-parameters of ML models trained by ML service provider~\cite{wang2018stealing}. 

In opposition to above attacks, membership encoding aims to encode and infer the exact membership information for a subset of  training set rather than some properties of a class or a subgroup in the training data.




%% file: conclusion.tex
\section{Conclusions}
We presented membership encoding that forces deep learning model to learn distinguishable representation for a substantial fraction of training data with negligible loss in test accuracy. Our experimental results showed that one can perform accurate membership inference on target data records even if the input data is redacted or the trained model is being modified. Our encoding mechanism can be used for watermarking for copyright protection as well as inferring membership of sensitive training data.